\documentclass[letterpaper, 10 pt, journal, twoside]{IEEEtran}

\IEEEoverridecommandlockouts                              

\usepackage{cite}
\usepackage{amsmath,amssymb,amsfonts}
\usepackage{algorithmic}
\usepackage{graphicx}
\usepackage{textcomp}
\usepackage{xcolor}
\usepackage{color}
\usepackage{colortbl}
\usepackage{mathtools}
\usepackage{amsmath}
\usepackage{tabularx}
\usepackage{multirow}
\usepackage{multicol}
\usepackage{float}
\usepackage{tabularx}
\usepackage{threeparttable}
\usepackage{booktabs}
\usepackage{longtable}
\usepackage{makecell}
\usepackage{verbatim}
\usepackage{wrapfig}
\usepackage{xspace}

\definecolor{SA}{rgb}{1.0,0.92,0.84}
\definecolor{CA}{rgb}{0.92,0.96,1.0}
\definecolor{BT}{rgb}{0.89,0.98,0.89}
\definecolor{red}{rgb}{0.0,0.0,0.0}

\newcommand{\model}{TEG-Track\xspace}
\def\BibTeX{{\rm B\kern-.05em{\sc i\kern-.025em b}\kern-.08em
    T\kern-.1667em\lower.7ex\hbox{E}\kern-.125emX}}

\begin{document}

\title{
Enhancing Generalizable 6D Pose Tracking \\of an In-Hand Object with Tactile Sensing
}

\author{Yun Liu$^{*,1,2}$, Xiaomeng Xu$^{*,1}$, Weihang Chen$^{3}$, Haocheng Yuan$^{4}$, He Wang$^{5}$, Jing Xu$^{3}$, Rui Chen$^{3}$, and Li Yi$^{1,2,6}$
\thanks{Manuscript received: July, 18, 2023; Revised October, 17, 2023; Accepted November, 20, 2023.}
\thanks{This paper was recommended for publication by Editor Pascal Vasseur upon evaluation of the Associate Editor and Reviewers’ comments.}
\thanks{Project supported by the Young Scientists Fund of the National Natural Science Foundation of China (Grant No. 62203258).}
\thanks{$^*$Yun Liu and Xiaomeng Xu are co-first authors.}
\thanks{Li Yi is the corresponding author.}
\thanks{$^1$Institute for Interdisciplinary Information Sciences, Tsinghua University, Beijing, China}
\thanks{$^2$Shanghai Qizhi Institute, Shanghai, China}
\thanks{$^3$Department of Mechanical Engineering, Tsinghua University, Beijing, China}
\thanks{$^4$Northwestern Polytechnical University, Xian, China}
\thanks{$^5$Center on Frontiers of Computing Studies, Peking University, Beijing, China}
\thanks{$^6$Shanghai AI Laboratory, Shanghai, China}
\thanks{Digital Object Identifier (DOI): see top of this page.}
}

\markboth{IEEE Robotics and Automation Letters. Preprint Version. Accepted November, 2023}
{Liu \MakeLowercase{\textit{et al.}}: Enhancing Generalizable 6D Pose Tracking of an In-Hand Object with Tactile Sensing}

\maketitle
\thispagestyle{empty}
\pagestyle{empty}

\begin{abstract}

When manipulating an object to accomplish complex tasks, humans rely on both vision and touch to keep track of the object's 6D pose. However, most existing object pose tracking systems in robotics rely exclusively on visual signals, which hinder a robot's ability to manipulate objects effectively. To address this limitation, we introduce \model, a tactile-enhanced 6D pose tracking system that can track previously unseen objects held in hand.  From consecutive tactile signals, \model optimizes object velocities from marker flows when slippage does not occur, or regresses velocities using a slippage estimation network when slippage is detected. The estimated object velocities are integrated into a geometric-kinematic optimization scheme to enhance existing visual pose trackers. To evaluate our method and to facilitate future research, we construct a real-world dataset for visual-tactile in-hand object pose tracking. Experimental results demonstrate that \model consistently enhances state-of-the-art generalizable 6D pose trackers in synthetic and real-world scenarios. \textit{Our code and dataset are available at https://github.com/leolyliu/TEG-Track.}

\end{abstract}



\section{INTRODUCTION}
\begin{figure}[!tbp]
    \centering
    \vspace{0.25cm}
    \includegraphics[width=0.85\columnwidth]{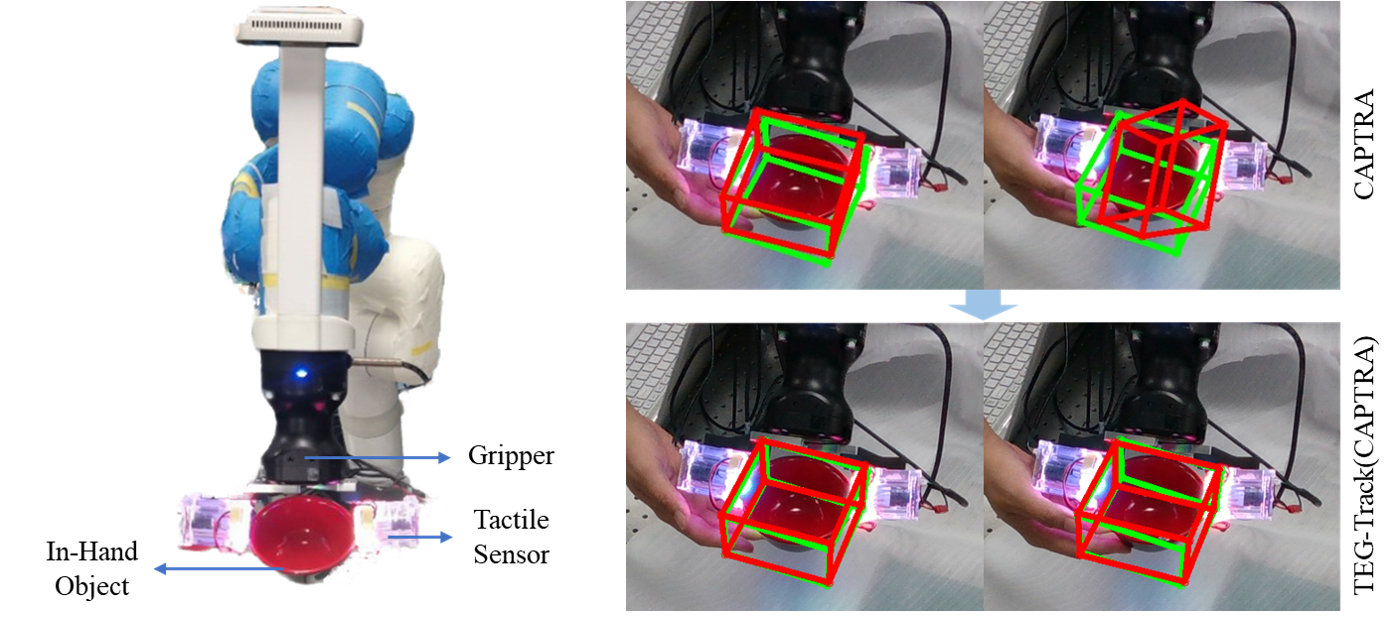}
    \vspace{-0.4cm}
    \caption{We propose a general in-hand object pose tracking framework \model, and evaluate it on our synthetic and real datasets. Our approach enhances generalizable visual trackers such as BundleTrack\cite{wen2021bundletrack} with tactile sensing. Here we visualize the tracking task as tracking the object's 3D bounding box: green boxes denote ground truth poses whereas red boxes denote estimated poses.}
     \vspace{-0.4cm}
    \label{fig:task_intro}
\end{figure}

\IEEEPARstart{A}{ccurate} 6D pose tracking of objects is essential for enabling effective robotic manipulation. Prior research\textcolor{red}{~\cite{xiang2018posecnn, He_2020_CVPR, wen2020se}} has demonstrated impressive precision and robustness for tracking known objects using 3D object models. Recent studies have \textcolor{red}{further} shifted their focus towards developing generalizable 6D pose tracking methods that can handle novel object instances from known~\cite{wang20206,weng2021captra,lin2022keypoint} or even unknown\textcolor{red}{~\cite{wen2021bundletrack,nguyen2022pizza,wen2023bundlesdf}} object categories. In this paper, we contribute to the development of such generalizable 6D pose tracking techniques, specifically addressing the \textbf{in-hand} setup shown in Figure \ref{fig:task_intro} that is commonly encountered in robot manipulation tasks.
Our goal is to consecutively track the 6D pose of an in-hand object starting from its initial 6D pose.
In scenarios where objects are manipulated by robot hands, relying solely on robot proprioception could prove challenging, particularly when external forces from collisions or multi-agent interactions occur. Therefore, it is critical to have an accurate in-hand object tracker that can precisely capture the object's state, especially in contexts involving in-hand manipulation or rich environmental contacts such as peg-hole insertion. Furthermore, this research could significantly benefit human-robot collaboration \cite{yang2021reactive, laplaza2022context, ng2022takes}, where sudden changes in the object's kinematic state caused by interactions are common.

Existing generalizable 6D pose tracking methods face challenges in in-hand manipulation scenarios.
Compared with scenes without robot manipulation, visual sensing of the in-hand object become more distorted and less informative due to in-hand occlusions, which could impede existing methods that heavily rely on only \textcolor{red}{visual} signals like RGB-D images.
As a remedy, tactile sensing could be \textcolor{red}{integrated} into the tracking process. By equipping the robot hand with tactile sensors such as GelSight\cite{yuan2017gelsight}, we can capture high-quality geometric and motion signals from contact areas. Such \textcolor{red}{information from} tactile sensing \textcolor{red}{can} complement the noisy visual sensing caused by occlusions, meanwhile combining with the rapid advancement of tactile sensor technologies\cite{yuan2017gelsight, gelslim, digit, omnitact}, making the integration feasible and promising. Moreover, precise tactile sensing captures accurate motions for object contact regions, providing strong clues for understanding object pose changes.

Therefore, we propose \model, a general framework for enhancing generalizable 6D pose tracking of an in-hand object with tactile sensing.
First, from tactile sensing alone, \model learns tactile kinematic cues that indicate the kinematic states of the object.
Combining with visual sensing, \model then integrates object kinematic states with existing generalizable visual pose trackers through a geometric-kinematic optimization strategy.
\model can be easily plugged into various generalizable pose trackers, including template-based (introduced in Section \ref{sec:5.2}), regression-based\cite{weng2021captra}, and keypoint-based\cite{wen2021bundletrack} approaches.

To evaluate \model, we curate synthetic and real-world datasets due to the lack of datasets supporting generalizable visual-tactile in-hand object pose tracking research.
Since existing datasets\cite{dikhale2022visuotactile,tu2023posefusion} only serve for \textcolor{red}{single-frame} object pose estimation with a small data scale, we collect a large-scale synthetic object pose tracking dataset with large in-hand motion variations to test \model widely in various situations. Furthermore, to examine \model in real scenarios, we contribute a real-world visual-tactile in-hand object pose tracking dataset including 200 trajectories covering 17 instances from 5 object categories with careful per-frame object pose annotations. Experiments demonstrate that \model consistently improves the performances of different generalizable visual pose trackers in both synthetic and real scenarios. Compared \textcolor{red}{to} a state-of-the-art \textcolor{red}{generalizable} pose tracker BundleTrack\cite{wen2021bundletrack} on our real evaluation set, \model achieves 30.9$\%$ and 21.4$\%$ decreases in the average rotation and translation errors, respectively.

In summary, our main contributions are \textcolor{red}{threefold}:
1) \textcolor{red}{To the best of our knowledge, we} are \textcolor{red}{among} the first to explore generalizable in-hand object pose tracking combining visual and tactile sensing.
2) We present \model, a visual-tactile framework that learns tactile kinematic cues from tactile sensing and then incorporates them into various visual pose trackers with consistent performance gain.
3) \textbf{Dataset}: We construct the first fully-annotated visual-tactile in-hand object pose tracking dataset in real-world scenarios to facilitate future research.


\section{Related Work}
\textbf{Generalizable Visual Pose Tracking.}
Different from instance-level \textcolor{red}{object pose} tracking
\cite{deng2021poserbpf,wen2020se}
, generalizable \textcolor{red}{object pose tracking methods\cite{wang20206,weng2021captra,nguyen2022pizza,lin2022keypoint,wen2021bundletrack,wen2023bundlesdf} aim} to track \textcolor{red}{the pose for an unseen} object without its 3D model \textcolor{red}{and can be divided into regression-based and keypoint-based methods}.
Regression-based approaches\cite{weng2021captra,nguyen2022pizza} \textcolor{red}{directly use} a neural network to regress 6D object motion from RGB\cite{nguyen2022pizza} or point cloud\cite{weng2021captra} sequences, \textcolor{red}{while}
 keypoint-based methods\cite{wang20206,lin2022keypoint,wen2021bundletrack,wen2023bundlesdf} \textcolor{red}{are two-stage} that first detect object keypoints and then estimate \textcolor{red}{object pose differences among different frames by keypoint matching}.
\textcolor{red}{In terms of generalizability, category-level trackers\cite{weng2021captra,wang20206,lin2022keypoint} are limited to objects from known object categories during test time, while category-agnostic ones\cite{nguyen2022pizza,wen2021bundletrack,wen2023bundlesdf} can track an arbitrary object without the category information. However, visual signals are the only input for these methods, impeding them to apply to robot manipulation scenarios due to visually heavy occlusions.}

\textbf{Visual-Tactile 3D Perception and Datasets.} Tremendous efforts
\cite{smith20203d, wang2021elastic, wang20183d, smith2021active, suresh2022shapemap, chaudhury2022using, yang2023tacgnn, rustler2022active, rustler2023efficient} 
have been made to combine visual and tactile signals to deal with several 3D perception tasks other than pose tracking.
To reconstruct the 3D shape of the object in contact, a line of \textcolor{red}{studies}\cite{smith20203d, wang2021elastic, suresh2022shapemap, xu2023visual} \textcolor{red}{first reconstructs} a coarse \textcolor{red}{object mesh} by visual sensing \textcolor{red}{and then refines the details with tactile information}, and others\cite{wang20183d, smith2021active, rustler2022active, rustler2023efficient} further design iterative strategies to \textcolor{red}{online} search for \textcolor{red}{a} local \textcolor{red}{object region} \textcolor{red}{with} the most \textcolor{red}{informative} tactile signals.
To estimate the pose of a static object from an active robot movement, a multi-stage method\cite{chaudhury2022using} leverages visual and tactile sensing alternatively in different robot states.
Various visual-tactile datasets have been collected to facilitate studies \textcolor{red}{on} object shape reconstruction\cite{gao2021objectfolder,gao2022objectfolder}, in-hand object pose estimation\cite{dikhale2022visuotactile,tu2023posefusion}, and object grasping\cite{zhang2021partial,https://doi.org/10.48550/arxiv.2209.05022}.
We present the first \textcolor{red}{real-world} visual-tactile dataset \textcolor{red}{supporting researches on object pose tracking}.

\textbf{Object Pose Estimation and Tracking via Tactile Feedback.}
Due to the relatively low quality of visual sensing, previous works have explored object pose estimation and tracking via tactile feedback, but
limited to instance-level tracking with a major focus on static grasps.
\textcolor{red}{To estimate the object pose in a single frame, a tactile-only method\cite{gao2023hand} combines multiple tactile images via proprioception. Recent studies\cite{wen2020robust,https://doi.org/10.48550/arxiv.2301.13667} in this field combines visual and tactile sensing to achieve object pose estimation. \textit{Wen et.al.}\cite{wen2020robust} generates pose hypotheses from visual point clouds and then prunes them via hand-object collision check, and \textit{Caddeo et.al.}\cite{https://doi.org/10.48550/arxiv.2301.13667} encodes different modalities to learnable features and fuses them in the feature space.}
\textcolor{red}{To track the object pose with a known object model}, \textit{{\'A}lvarez et.al.}\cite{alvarez2019visual} separately predicts the object pose using visual and tactile signals alone, then fuses \textcolor{red}{the two predictions by an extended} Kalman Filter.
Another method\cite{izatt2017tracking} fuses visual \textcolor{red}{and} tactile point clouds \textcolor{red}{and then align the integral point cloud to the object model}.


\section{Method}
\begin{figure*}
    \centering
    \vspace{0.25cm}
    \includegraphics[width=0.75\textwidth]{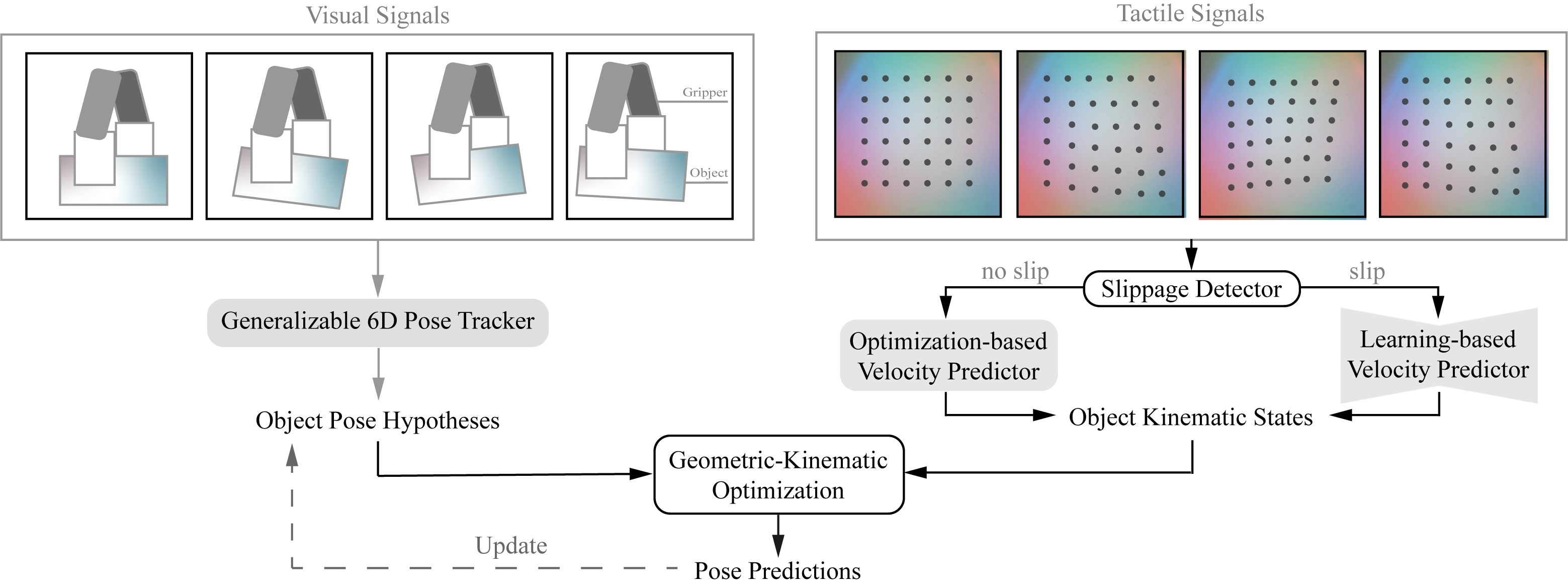}
    \vspace{-0.4cm}
    \caption{Overview of \model. During the online pose tracking process, TEG-Track first estimates object pose hypotheses by a generalizable visual pose tracker, meanwhile predicting the object's kinematic state from tactile kinematic cues learned from tactile signals. Incorporating kinematic states into pose hypotheses, \model then refines the object poses via a geometric-kinematic optimization strategy and utilizes final results to help subsequent tracking.
    }
    \vspace{-0.4cm}
    \label{fig:method_overview}
\end{figure*}

In this section, we introduce \model in detail.
As illustrated in Figure \ref{fig:method_overview}, the key idea is leveraging tactile kinematic cues learned from tactile sensing to boost visual pose trackers through a geometric-kinematic optimization strategy.
We first revisit generalizable visual pose trackers in Section \ref{sec:3.1}. We then present tactile kinematic cues that estimate kinematic state of the in-hand object from tactile images and marker flows in Section \ref{sec:3.2}. Finally, we propose a geometric-kinematic optimization strategy to integrate object kinematic states with various visual pose trackers in Section \ref{sec:3.3}.

\subsection{Generalizable Visual Pose Trackers}
\label{sec:3.1}
A generalizable pose tracker aims at transferring the learned pose tracking policy to novel objects without their 3D models. \textcolor{red}{For instance, CAPTRA\cite{weng2021captra} trains one network per object category, and uses it to track an unseen object from the same category during test time.} Such a pose tracker can \textcolor{red}{be designed} in a template-based, regression-based, or keypoint-based manner. A general object-centric representation (complete object model, NOCS Map\cite{wang2019normalized}, object keypoints, etc.) is commonly regarded as an intermediate object feature to build the bridge between visual inputs and 3D object pose.
Though generalizable visual pose trackers have achieved impressive results, their heavy reliance on visual sensing makes it difficult to handle heavy occlusion under in-hand situations. \model leverages a generalizable visual pose tracker to provide object pose hypotheses and then incorporates tactile sensing to improve them as final pose results.

\subsection{Tactile Kinematic Cues}
\label{sec:3.2}

\begin{figure}
\centering
\includegraphics[width=1.0\columnwidth]{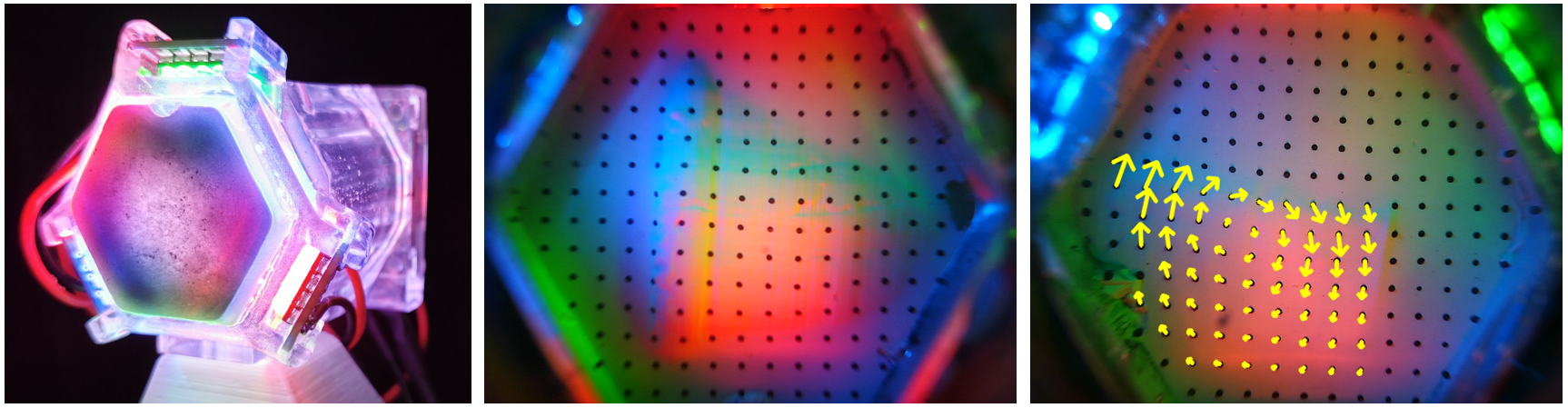}
\vspace{-0.7cm}
\caption{Tactile sensor, tactile RGB image and detected marker flows.}
\vspace{-0.5cm}
\label{fig:marker_motion}
\end{figure}

As shown in Figure \ref{fig:marker_motion}, when a robot manipulates an object, tactile RGB images capture high-quality geometry of the object's contact area. Differences between geometries in adjacent frames can indicate kinematic states of the in-hand object consisting of linear and angular velocities, which benefits the understanding of object pose changes. We thus present a learnable mapping from tactile images to object kinematic states, dubbed tactile kinematic cues. In general, such a mapping fulfills in a data-driven manner. Despite complex object motion caused by collisions and interactions, we observe that the mapping is simple when the object approximately sticks to the tactile sensor, thus presenting a two-stage design.
First, a slippage detector is used to detect whether the in-hand object is slipping between adjacent frames.
Then, object kinematic states are estimated by a velocity predictor via an optimization-based manner if no slippage occurs, otherwise via a learning-based network.

\textbf{Slippage Detector.} Given tactile signals from two adjacent frames, we determine if the in-hand object has slippage. As shown in Figure \ref{fig:marker_motion}, optical tactile sensors\cite{yuan2017gelsight,gelslim,taylor2022gelslim} could capture marker positions and further compute marker flows via Nearest-Neighbor matching. We draw inspiration from GelSlim3.0\cite{taylor2022gelslim} and detect object slippage by fitting an affine transformation to the marker flows between consecutive frames. Specifically, denoting $N_c$ as the number of marker flows, and $S=\{s_i \in \mathbb{R}^2\}_{i=1}^{N_c}$ and $T=\{t_i \in \mathbb{R}^2\}_{i=1}^{N_c}$ as source and target pixels of marker flows, we compute $E=\min\limits_{A,b} \sum_{i=1}^{N_c} \Vert A s_i + b - t_i \Vert_2^2$ via the least-square method. A slippage is detected if and only if $E$ is larger than a threshold.

\textbf{Optimization-based Velocity Predictor.} When a visual tactile sensor is in contact with a rigid object, 
the contact area of the sensor surface sticks to the grasped object and moves simultaneously if no slippage occurs.
In this case, the \textcolor{red}{2D pixel flow $f_{c_i}$} of a contact point $p_{c_i}$ on the sensor, which \textcolor{red}{is} approximate to the point on the object surface, can be acquired from the marker flows detected from consecutive tactile images exemplified in Figure \ref{fig:marker_motion}. \textcolor{red}{$f_{c_i}$ is then transformed to 3D velocity $v_{c_i}$ via tactile depth map estimated by GelSight\cite{yuan2017gelsight}.}

\textcolor{red}{As a rigid body, the object's motion can be defined as a linear velocity $v \in \mathbb{R}^3$ and an angular velocity $\omega \in \mathbb{R}^3$ at a pivot point $\hat{p} \in \mathbb{R}^3$ in the world coordinate system.} Given $N_c$ contact points detected from the tactile sensor \textcolor{red}{and a point $\hat{p}$}, we can optimize $v$ and $\omega$ from the detected positions $P_c = \{p_{c_i} \in \mathbb{R}^3\}_{i=1}^{N_c}$ and velocities $V_c = \{v_{c_i} \in \mathbb{R}^3\}_{i=1}^{N_c}$ of these contact points.
For each contact point \textcolor{red}{$p_{c_i}$}
we have $\textcolor{red}{v_{c_i}}=v+\omega \times \textcolor{red}{(p_{c_i}-\hat{p})}$. Based on such a constraint from each contact point, we define the energy function:

\vspace{-0.4cm}
\begin{equation}
\label{equa:kinematic}
E_{kinematics}(\hat{v}, \hat{\omega}) = \sum_{i=1}^{N_c} \Vert v_{c_i} - \hat{v} - \hat{\omega} \times (p_{c_i} - \textcolor{red}{\hat{p}}) \Vert ^ 2.
\end{equation}

\textcolor{red}{We set $\hat{p}$ to the estimated object position in the last frame. Then, velocity estimates $\hat{v}$ and $\hat{\omega}$} are computed by minimizing $E_{kinematics}$ \textcolor{red}{via the least-square method}. \textcolor{red}{The estimated object kinematic state is formulated as $\{\hat{v},\hat{\omega},\hat{p}\}$ and will be used in our geometric-kinematic optimization.}

\textbf{Learning-based Velocity Predictor.} When the in-hand object slips on the sensor surface, the contact regions on the object change rapidly, making contact point information of tactile sensors unreliable. We thus design a learning-based approach to directly regress object kinematic states from raw adjacent tactile images without the usage of marker points.

\begin{figure}[H]
\centering
\includegraphics[width=0.9\columnwidth]{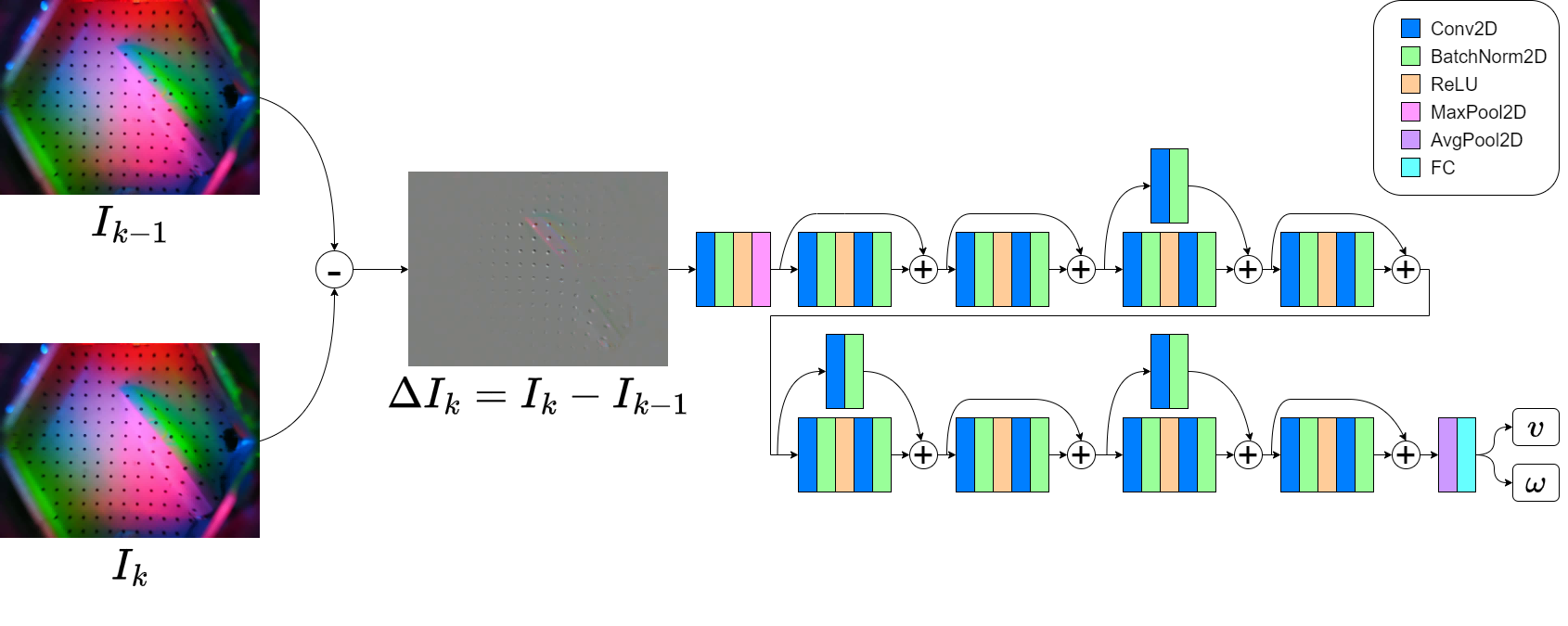}
\vspace{-0.4cm}
\caption{\textcolor{red}{The network structure of our learning-based velocity predictor.}}
\vspace{-0.2cm}
\label{fig:learning_VP}
\end{figure}

\textcolor{red}{Figure \ref{fig:learning_VP} illustrates our learning-based velocity predictor.} Given the difference $\Delta I_k$=$I_k$-$I_{k-1} \in \mathbb{R}^{384\times288\times3}$ between two adjacent tactile RGB images $\{I_{k-1}, I_k\}$ as input, our network first encodes \textcolor{red}{it to} a 512D feature \textcolor{red}{vector} using a ResNet\cite{he2016deep} structure, then decodes \textcolor{red}{the vector} to a 6D velocity representation $[v, \omega]$ via a fully-connected layer. \textcolor{red}{$v$ and $\omega$ from the network are defined at the origin of the robot gripper's coordinate system, and finally transformed to velocities at the origin of the world coordinate system by applying known robot gripper pose.}
\textcolor{red}{The loss function $\mathcal{L} = \lambda_v \Vert v - \tilde{v} \Vert _2^2 + \lambda_\omega \Vert \omega - \tilde{\omega} \Vert _2^2$ is the weighted sum of mean-square errors on $v$ and $\omega$,
where $\tilde{v}$ and $\tilde{\omega}$ are ground-truth velocities. We set $\lambda_v=100$ and $\lambda_\omega=0.25$. The network is train by an Adam optimizer with learning rate $1e-4$.}

\subsection{Integrating Object Kinematic States with Generalizable Vision-based Pose Trackers}
\label{sec:3.3}

Given the estimated kinematic state of the in-hand object, a straightforward kinematics-only tracking solution is to compute the object pose differences between adjacent frames with object velocities and time differences. However, since velocity information lacks rectification of object pose bias, using such kinematic states alone will yield significant accumulation error. Meanwhile, visual estimate alone has a relatively small accumulation error and maintains a stable error range, while it fluctuates due to heavy sensor noise. We observe that visual and tactile modalities are complementary to achieve a better tracking performance, and thus propose a geometric-kinematic optimization strategy that enhances geometry-based object pose hypotheses from generalizable visual trackers with the estimated kinematic states.
    
\textbf{Geometric-Kinematic Optimization.}
For the $i$-th frame, the object state is defined as $\{t_i,r_i,v_i,\omega_i, \textcolor{red}{p_i}\}$ including position $t_i\in \mathbb{R}^3$, orientation $r_i\in \mathbb{R}^3$, linear velocity $v_i\in \mathbb{R}^3$ and angular velocity $\omega_i\in \mathbb{R}^3$ \textcolor{red}{that are defined at pivot point $p_i \in \mathbb{R}^3$}.
\textcolor{red}{Object kinematic state prediction} $\{\hat{v}_i, \hat{\omega}_i, \textcolor{red}{\hat{p}_i}\}$ \textcolor{red}{is} provided by velocity predictors in Section \ref{sec:3.2}.
During the online pose tracking process, when handling the $k$-th frame, we first obtain vision-based object pose hypothesis $\{\bar{t}_k, \bar{r}_k\}$ for frame $k$ from Section \ref{sec:3.1}, then incorporate object kinematic states $\{\hat{v}_i,\hat{\omega}_i\textcolor{red}{,\hat{p}_i}\}$ into pose hypotheses $\{\bar{t}_i, \bar{r}_i\}$ for frames $[k$-$N$+$1,k]$ to predict current object pose $\{t_k, r_k\}$, where $N$ is a manually-designed parameter.
Given a guess $\{\hat{t}_k, \hat{r}_k\}$ for $\{t_k, r_k\}$, the pose predictions $\{\hat{t}_i, \hat{r}_i\}(i\in [k$-$N$+$1,k$-$1])$ can be recursively computed by:

\vspace{-0.4cm}
\begin{equation}
\label{euqa:delta_pose}
\begin{split}
& \hat{t}_{i}=\hat{t}_{i+1}-\textcolor{red}{(\hat{v}_{i+1} + \hat{\omega}_{i+1} \times \overrightarrow{\hat{t}_{i+1}-\hat{p}_{i+1}})}\cdot \Delta T_{i+1},\\
& \hat{r}_{i}=\{I+\sin(\Vert \hat{\omega}_{i+1} \Vert \cdot \Delta T_{i+1})\cdot [\frac{-\hat{\omega}_{i+1}}{\Vert \hat{\omega}_{i+1} \Vert}\times]+\\
& [1-\cos(\Vert \hat{\omega}_{i+1} \Vert \cdot \Delta T_{i+1})] \cdot [\frac{-\hat{\omega}_{i+1}}{\Vert \hat{\omega}_{i+1} \Vert} \times]^2\} \cdot \hat{r}_{i+1},
\end{split}
\end{equation}

where $\Delta T$ indicates the time interval between adjacent frames, $[\omega\times]$ denotes the skew-symmetric matrix of $\omega$.

We use $E_t(\hat{t}, \bar{t})$=$\Vert \hat{t}$-$\bar{t} \Vert_2^2$
to measure the distance between two positions, and \textcolor{red}{$E_r(\hat{r}, \bar{r})$=$\Vert R(\hat{r})$-$R(\bar{r}) \Vert_2^2$} to measure the \textcolor{red}{difference} between two orientations, where $R(r)$ indicates the rotation matrix of orientation $r$.
Using pose hypotheses $\{\bar{t}_i, \bar{r}_i\}$ as constraints, the final pose predictions \textcolor{red}{$\{\hat{t}_k, \hat{r}_k\}$} are estimated by minimizing the following energy function via a multi-frame optimization procedure:

\vspace{-0.4cm}
\begin{equation}
\label{equa:tracking}
E_{track}(\hat{t}_k, \hat{r}_k)=\sum_{i=k-N+1}^{k}E_t(\hat{t}_i, \bar{t}_i) + E_r(\hat{r}_i,\bar{r}_i),
\end{equation}

where $\{\hat{t}_i, \hat{r}_i\}(i<k)$ are computed by Equation \ref{euqa:delta_pose}.
We use Adam \cite{kingma2017adam} optimizer to find $\{\hat{t}_k^*,\hat{r}_k^*\}$ that minimizes Equation \ref{equa:tracking}, and then obtain $\{\hat{t}_i^*, \hat{r}_i^*\}$ ($i\in[k$-$N$+$1,k]$) via Equation \ref{euqa:delta_pose} as our final pose predictions.
The final results are used to update object pose hypotheses for the following tracking procedure.


\section{Dataset}
\begin{figure}
    \centering
    \vspace{0.25cm}
    \includegraphics[width=0.95\columnwidth]{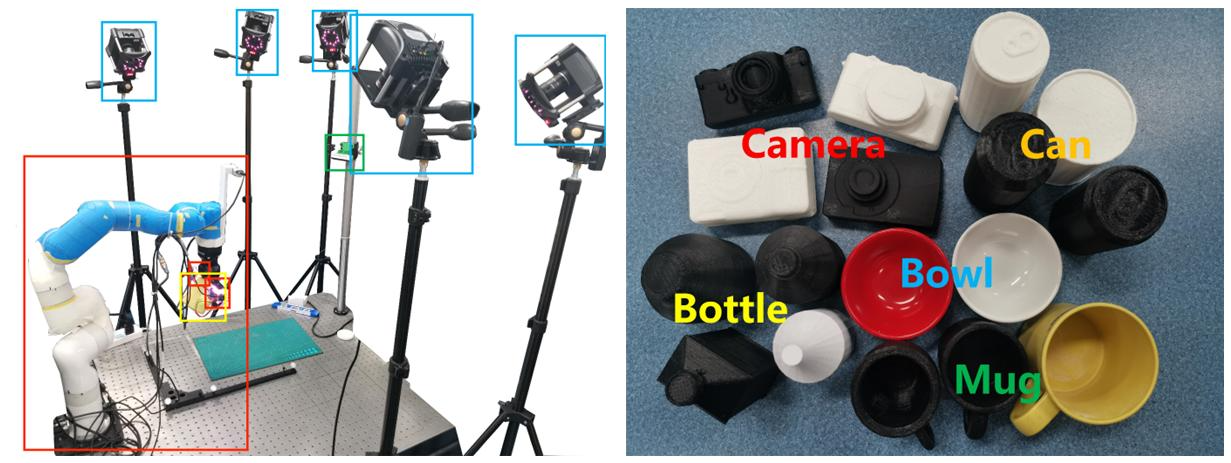}
    \vspace{-0.3cm}
    \caption{(a) The data capturing system. The red, green, blue and yellow boxes indicate the Xarm robot arm with Geisight tactile sensors, the Realsense D415 RGB-D sensor, the NOKOV motion capture suite and the in-hand tracked object, respectively. (b) Object categories and instances.}
    \vspace{-0.5cm}
    \label{fig:dataset_configure}
\end{figure}

The hardware devices of both synthetic and real-world data capturing systems include a Xarm7 robot arm, a Robotiq 2F-140 gripper, a third-view RealSense D415 camera, and two GelSight\cite{yuan2017gelsight} tactile sensors mounted on the gripper. As illustrated in Figure \ref{fig:dataset_configure}, in our real-world scenario, we additionally set up a NOKOV motion capture system to obtain the ground-truth object poses. The simulation and real-world settings share similar hardware parameters.

\subsection{Synthetic Dataset}
We carefully select five object categories (camera, can, bottle, mug, bowl) from NOCS\cite{wang2019normalized} that are easy to hold by robot gripper and incorporate two (earphone, birdhouse) to enrich the variance of object geometry. All objects are chosen from ShapeNet\cite{chang2015shapenet} dataset. We establish simulation in the SAPIEN\cite{xiang2020sapien} environment, place object collision meshes on the table plane and leverage motion planning to control the robot arm to automatically lift the object, move it, \textcolor{red}{and finally push it to contact the table}. For each object category, we collect 10-41 object instances with different geometries and 90-277 successful robot manipulation trajectories. We follow a ratio of 7:3 to randomly separate the objects into training and evaluation sets. The data capturing frequency of robot manipulation is 30 FPS. Each trajectory is approximately 100 frames in total.
For each video frame, we capture a visual RGB-D image, tactile RGB images from two tactile sensors mounted on each robot finger, tactile depth images reconstructed via rendering, the position and velocity of each sensor contact point, a 2D object mask, and the object pose directly obtained by the simulation environment.
To minimize the gap between simulation and the real world, we model visual and tactile sensor noise following the pipelines in \cite{zhang2023close}, \cite{9756938}, \cite{yuan2017gelsight}, and empirically add multiple Gaussian noises to the positions and linear velocities of contact points based on the sensor noise patterns observed in the real world.

\subsection{Real Dataset}
We construct the first real-world visual-tactile in-hand object pose tracking dataset. We believe this is strong support for future relevant research efforts.

\begin{figure}
    \centering
    \vspace{0.25cm}
    \includegraphics[width=0.95\columnwidth]{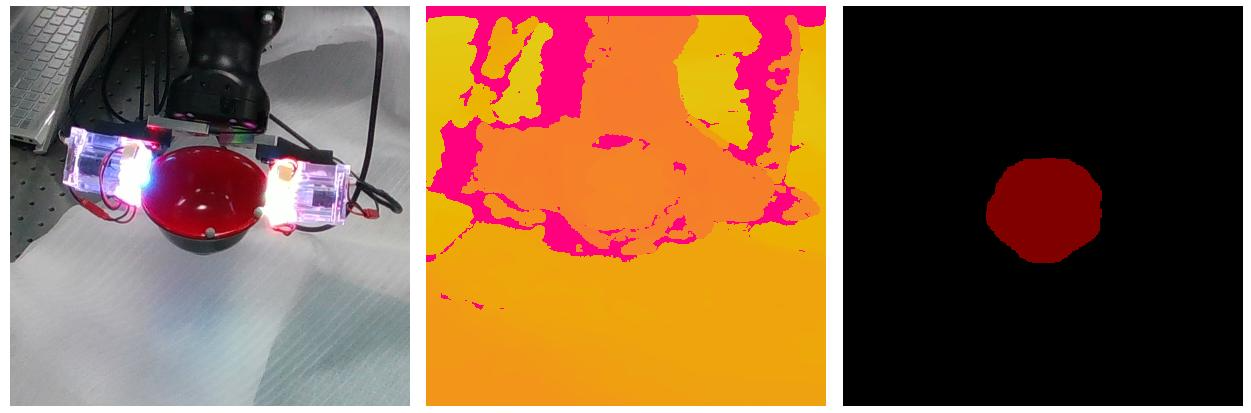}
    \vspace{-0.4cm}
    \caption{Visual signals in our real-world dataset: RGB image, depth image, 2D object mask.}
    \vspace{-0.4cm}
    \label{fig:real_visual_signals}
\end{figure}

We collect 200 videos among 17 object instances (shown in Figure \ref{fig:dataset_configure}(b)) spanning five object categories (camera, can, bottle, mug, and bowl). The integrated system captures the visual-tactile signals and the object pose at approximately 20 FPS, and each video contains 401 RGB-D frames. Besides the object poses captured by the motion capture system, we also provide the 2D masks of the in-hand object via a semi-automatic annotation process using the MIVOS\cite{mivos} tool. The sorts of visual signals are shown in Figure \ref{fig:real_visual_signals}. We follow a ratio of 3:7 to construct training and evaluation sets, respectively. The training set \textcolor{red}{includes three categories (camera, can, bottle) with 6 objects that are not included in the test set, which} is only used to train the learning-based velocity predictor in our experiments.

In real scenarios, we manually apply various motions to the in-hand object: 1) We impose collisions between the manipulated object and background objects. The background object held in the human hand touches the manipulated object from multiple positions and directions, which enriches the collision patterns and significantly increases the difficulty of pose tracking. 2) The object could be touched by human hands and hence produce a fast in-hand movement which is smoother than that in 1). 3) We apply zero-force control on the robot arm. The robot arm is moved by manually dragging, hence the robot motion is rich and smooth and could keep the tracking challenging.
Table \ref{tab:real_motion} shows \textcolor{red}{the average object speed} in adjacent video frames, indicating \textcolor{red}{the object motion is generally fast with a large variance}.
\begin{table}
    \centering
    \vspace{0.25cm}
    \caption{AVERAGE MOTION OF THE IN-HAND OBJECT.}
    \vspace{-0.25cm}
    \begin{tabular}{|c|c|c|c|}
        \hline
        Category & Number of Video & \textcolor{red}{AV$^{\rm 1}$ ($^{\circ}$/s)} & \textcolor{red}{LV$^{\rm 2}$ (cm/s)} \\
        \hline
        Camera & 44 & \textcolor{red}{$9.4(\pm10.4)$} & \textcolor{red}{$1.5(\pm1.7)$} \\
        \hline
        Can & 44 & \textcolor{red}{$8.6(\pm11.2)$} & \textcolor{red}{$1.5(\pm1.9)$} \\
        \hline
        Bottle & 43 & \textcolor{red}{$7.2(\pm9.4)$} & \textcolor{red}{$1.5(\pm2.1)$} \\
        \hline
        Mug & 36 & \textcolor{red}{$8.0(\pm9.4)$} & \textcolor{red}{$1.2(\pm1.6)$} \\
        \hline
        Bowl & 33 & \textcolor{red}{$9.4(\pm13.4)$} & \textcolor{red}{$0.8(\pm1.1)$} \\
        \hline
    \end{tabular}
    \label{tab:real_motion}
    \vspace{0.05cm}
    \\
    \footnotesize{\textcolor{red}{$^{\rm 1}$Angular velocity. $^{\rm 2}$Linear velocity.}
    }
    \vspace{-0.2cm}
\end{table}


\section{Experiments}
This section presents a comprehensive evaluation of \model.
We first introduce different baseline approaches including a tactile tracking method (Section \ref{sec:5.1}) and various vision-based generalizable pose trackers (Section \ref{sec:5.2}). We then compare \model with baseline approaches on both synthetic (Section \ref{sec:5.3}) and real-world (Section \ref{sec:5.4}) datasets. Additional ablation studies are presented in Section \ref{sec:5.5} to examine our method design. The tracking speed and robustness of \model are evaluated in Sections \ref{sec:5.6} and \ref{sec:5.7}, respectively. For \model, parameter $N$ in the geometric-kinematic optimization is set to 5.

\subsection{Kinematics-only Pose Tracking}
\label{sec:5.1}

Leveraging the kinematic cues from tactile sensors, object velocities $\{v, \omega\}$ can be estimated at the speed of 63 FPS on average via kinematic optimization and slippage estimation. Based on $\{v, \omega\}$, we can compute the difference of object pose between adjacent frames by Equation \ref{equa:kinematic}. Adding this pose difference to the last pose estimation could directly predict the current pose, hence a continuous pose tracking procedure can be achieved. This method is regarded as a tactile baseline approach in the following experiments.

\subsection{Different Choices of Vision-based Pose Trackers}
\label{sec:5.2}

We incorporate \model with three generalizable pose trackers: 1) ShapeAlign, a template-based category-level pose tracking method designed by our own, which registers visual point clouds to an estimated object model. ShapeAlign first acquires a category-level complete shape template of the object using PoinTr\cite{yu2021pointr}, then predicts object poses by aligning the shape template to visual point clouds via a Chamfer Distance loss. 2) CAPTRA\cite{weng2021captra}, a regression-based category-level pose tracking method. 3) BundleTrack\cite{wen2021bundletrack}, a keypoint-based category-agnostic pose tracking method for novel objects. ShapeAlign and CAPTRA take object depth point clouds alone as visual inputs, while BundleTrack utilizes RGB-D images and object masks.

\subsection{Results on Synthetic Data}
\label{sec:5.3}
Table \ref{tab:synthetic_result} summarizes the quantitative results for in-hand object pose tracking on seven object categories from our synthetic dataset, where K denotes the kinematics-only tracking method described in Section \ref{sec:5.1}, TE denotes our approach \model, SA denotes ShapeAlign, CA denotes CAPTRA, and BT denotes BundleTrack.

We report the following metrics: \textcolor{red}{1)} $5^{\circ}5$mm (\%), the percentage of pose estimation with translation error $\leq 5$mm and rotation error $\leq 5^{\circ}$. 
\textcolor{red}{2)} $R_{e}$ ($^{\circ}$), average rotation error. \textcolor{red}{3)} $T_{e}$ (mm), average translation error.

\model performs better than kinematics-only and vision-based methods on all \textcolor{red}{three} metrics, meanwhile bringing consistent improvements to three different types of generalizable visual pose trackers. For example, \model respectively brings an increment of 2.3$\%$, 13.3$\%$, and 2.6$\%$ to ShapeAlign, CAPTRA, and BundleTrack on the metric $5^{\circ}5$mm.

\begin{table}[t]
\setlength\tabcolsep{1.3pt}
\centering
\small
\vspace{0.25cm}
\caption{QUANTITATIVE RESULTS ON SYNTHETIC DATASET.}
\vspace{-0.25cm}
\begin{tabular}{|c|c|c|cc|cc|cc|}
    \hline
    \multicolumn{2}{|c|}{Method} & K & SA & TE(SA) & CA & TE(CA) & BT & TE(BT) \\
    \hline
    \multicolumn{2}{|c|}{Visual Signal} & N/A & \multicolumn{4}{c|}{Depth} & \multicolumn{2}{c|}{RGB-D} \\
    \hline
    \multirow{3}{*}{Camera} & $5^{\circ}5$mm$\uparrow$ & 9.8 & 42.3 & \textbf{55.3} & 53.5 & \textbf{60.8} & 87.1 & \textbf{89.8} \\
    & $R_{e}$($^{\circ}$)$\downarrow$ & 4.2 & 8.1 & \textbf{6.1} & 3.4 & \textbf{2.6} & 3.0 & \textbf{2.7} \\
    & $T_{e}$(mm)$\downarrow$ & 9.4 & 4.3 & \textbf{4.0} & 5.2 & \textbf{4.2} & 2.5 & \textbf{1.9} \\
    \hline
    \multirow{3}{*}{Can} & $5^{\circ}5$mm$\uparrow$ & 6.9 & \textbf{88.1} & 81.1 & 84.6 & \textbf{96.9} & \textbf{98.8} & 98.7 \\
    & $R_{e}$($^{\circ}$)$\downarrow$ & 7.0 & \textbf{2.3} & 2.4 & 1.3 & \textbf{0.9} & 1.4 & \textbf{1.2} \\
    & $T_{e}$(mm)$\downarrow$ & 10.5 & \textbf{1.8} & 2.7 & 3.8 & \textbf{2.0} & \textbf{1.4} & \textbf{1.4} \\
    \hline
    \multirow{3}{*}{Bottle} & $5^{\circ}5$mm$\uparrow$ & 5.0 & 50.8 & \textbf{55.2} & 61.2 & \textbf{80.3} & 95.5 & \textbf{96.1} \\
    & $R_{e}$($^{\circ}$)$\downarrow$ & 6.6 & 5.3 & \textbf{4.8} & 4.8 & \textbf{2.7} & 4.2 & \textbf{1.3} \\
    & $T_{e}$(mm)$\downarrow$ & 10.6 & \textbf{5.3} & 5.8 & 5.1 & \textbf{3.6} & 9.6 & \textbf{4.6} \\
    \hline
    \multirow{3}{*}{Earphone} & $5^{\circ}5$mm$\uparrow$ & 8.4 & 41.5 & \textbf{48.2} & 49.0 & \textbf{56.7} & \textbf{67.4} & 60.7 \\
    & $R_{e}$($^{\circ}$)$\downarrow$ & 3.5 & 10.9 & \textbf{6.5} & 5.4 & \textbf{4.2} & 4.5 & \textbf{4.4} \\
    & $T_{e}$(mm)$\downarrow$ & 9.6 & \textbf{4.1} & 4.2 & 3.3 & \textbf{2.4} & 2.0 & \textbf{1.2} \\
    \hline
    \multirow{3}{*}{Mug} & $5^{\circ}5$mm$\uparrow$ & 13.4 & 32.6 & \textbf{33.3} & 45.0 & \textbf{58.2} & 72.4 & \textbf{76.9} \\
    & $R_{e}$($^{\circ}$)$\downarrow$ & 3.0 & 16.0 & \textbf{7.9} & 7.1 & \textbf{5.5} & 5.1 & \textbf{4.0} \\
    & $T_{e}$(mm)$\downarrow$ & 9.0 & 4.5 & \textbf{4.4} & 4.0 & \textbf{3.1} & 2.7 & \textbf{2.2} \\
    \hline
    \multirow{3}{*}{Birdhouse} & $5^{\circ}5$mm$\uparrow$ & 7.0 & 29.3 & \textbf{34.6} & 42.9 & \textbf{50.1} & 72.8 & \textbf{80.3} \\
    & $R_{e}$($^{\circ}$)$\downarrow$ & 3.1 & 9.4 & \textbf{6.3} & 6.9 & \textbf{5.3} & 6.9 & \textbf{3.3} \\
    & $T_{e}$(mm)$\downarrow$ & 10.4 & \textbf{5.4} & 5.8 & 5.8 & \textbf{5.1} & 5.6 & \textbf{3.4} \\
    \hline
    \multirow{3}{*}{Bowl} & $5^{\circ}5$mm$\uparrow$ & 2.7 & \textbf{40.0} & 32.9 & 20.9 & \textbf{47.3} & 48.8 & \textbf{58.9} \\
    & $R_{e}$($^{\circ}$)$\downarrow$ & 12.5 & \textbf{8.3} & 12.0 & 19.3 & \textbf{4.2} & 9.3 & \textbf{6.6} \\
    & $T_{e}$(mm)$\downarrow$ & 11.7 & \textbf{2.1} & 4.2 & 7.2 & \textbf{4.8} & 7.5 & \textbf{1.5} \\
    \hline
    \multirow{3}{*}{Overall} & $5^{\circ}5$mm$\uparrow$ & 7.6 & 46.4 & \textbf{48.7} & 51.0 & \textbf{64.3} & 77.6 & \textbf{80.2} \\
    & $R_{e}$($^{\circ}$)$\downarrow$ & 5.7 & 8.6 & \textbf{6.6} & 6.9 & \textbf{3.6} & 4.9 & \textbf{3.4} \\
    & $T_{e}$(mm)$\downarrow$ & 10.2 & \textbf{3.9} & 4.4 & 4.9 & \textbf{3.6} & 4.5 & \textbf{2.3} \\
    \hline
\end{tabular}
\vspace{-0.3cm}
\label{tab:synthetic_result}
\end{table}

\subsection{Results on Real Data}
\label{sec:5.4}

Table \ref{tab:real_result} shows the quantitative results on our real dataset with the same metrics in Section \ref{sec:5.3}. Most experimental settings are the same as those on the synthetic dataset, except that CAPTRA is trained on synthetic data with the same object category for the lack of large-scale training data in real world. Consistent with the evaluation results on synthetic data, \model consistently enhances the performance of different vision-based pose trackers in real-world scenarios. Compared with ShapeAlign, \model improves learning-based CAPTRA and BundleTrack more significantly. One reason is that the instability of historical pose bias in the learning-based pose predictors could be alleviated by the geometric-kinematic optimization that integrates the information from multiple frames. The cumulative error of object pose severely influence the effect of the kinematics-only method, while it is mitigated by \model via the fusion with object visual information.

\begin{figure}[!tbp]
\centering

\includegraphics[width=0.85\columnwidth]{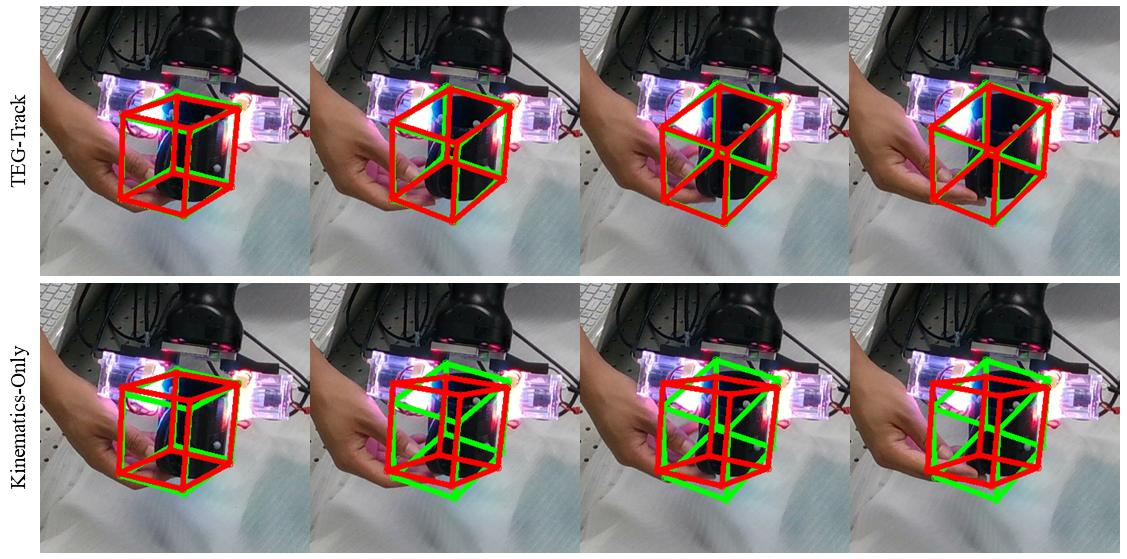}
\vspace{-0.4cm}
  \caption{Qualitative results of long-range trajectories on real data. Red and green bounding boxes indicate the predicted and the ground-truth poses of the in-hand object, respectively.
  }
  \vspace{-0.3cm}
  \label{fig:fig_real}
\end{figure}

\begin{figure}[!tbp]
\centering
\includegraphics[width=0.85\columnwidth]{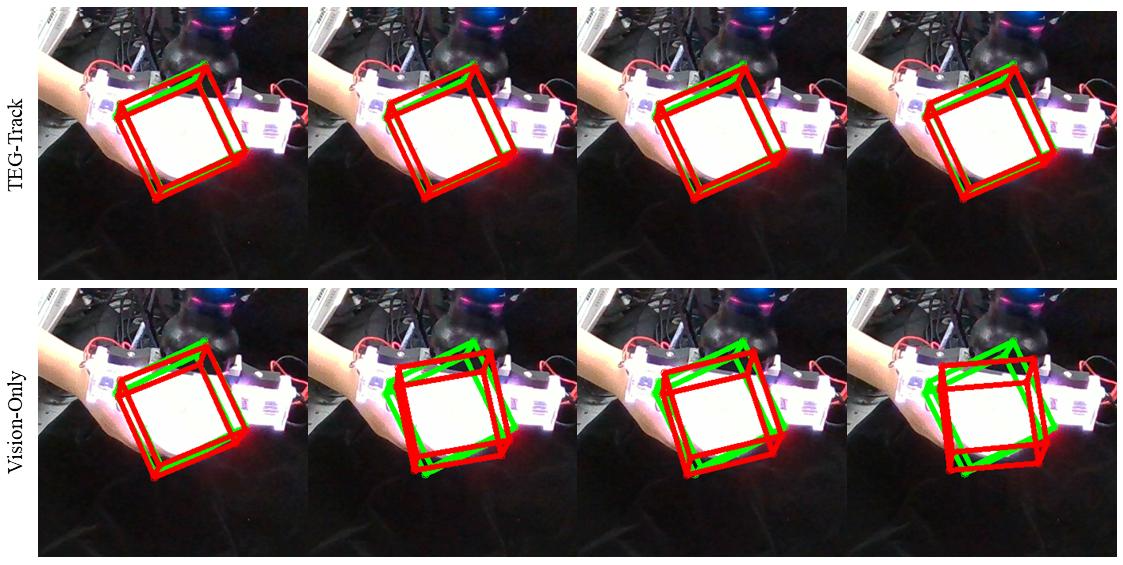}
\vspace{-0.4cm}
  \caption{Qualitative results of consecutive frames on real data.}
   \vspace{-0.5cm}
  \label{fig:fig_real_consecutive}
\end{figure}

The combination of visual and tactile signals benefits \model to track accurately and robustly, which helps \model perform better than the trackers that only leverage a single data modality. Figure \ref{fig:fig_real} shows the object pose predictions of \model(BundleTrack) at the 50, 150, 250, and 350-th frame of a real-world video, indicating that \model yields more accurate and stable results compared with the kinematics-only tracking method throughout a long period. Figure \ref{fig:fig_real_consecutive} shows the tracking results of \model(BundleTrack) in four adjacent frames, indicating \model can track more stably and precisely than visual pose trackers with the help of tactile kinematic cues.

\begin{table}[t]
\setlength\tabcolsep{1.3pt}
\centering
\small
\vspace{0.25cm}
\caption{QUANTITATIVE RESULTS ON REAL DATASET.}
\vspace{-0.25cm}
\begin{tabular}{|c|c|c|cc|cc|cc|}
    \hline
    \multicolumn{2}{|c|}{Method} & K & SA & TE(SA) & CA & TE(CA) & BT & TE(BT) \\
    \hline
    \multicolumn{2}{|c|}{Visual Signal} & N/A & \multicolumn{4}{c|}{Depth} & \multicolumn{2}{c|}{RGB-D} \\
    \hline
    \multirow{3}{*}{Camera} & $5^{\circ}5$mm$\uparrow$ & 76.9 & 66.5 & \textbf{69.8} & 0.8 & \textbf{1.4} & 85.2 & \textbf{88.8} \\
    & $R_{e}$($^{\circ}$)$\downarrow$ & 3.0 & 4.1 & \textbf{4.0} & 10.1 & \textbf{6.4} & 3.0 & \textbf{2.8} \\
    & $T_{e}$(mm)$\downarrow$ & 3.3 & 3.4 & \textbf{3.2} & 14.8 & \textbf{11.3} & 2.4 & \textbf{2.1} \\
    \hline
    \multirow{3}{*}{Can} & $5^{\circ}5$mm$\uparrow$ & 82.0 & 76.4 & \textbf{78.2} & 5.1 & \textbf{5.6} & 85.0 & \textbf{86.3} \\
    & $R_{e}$($^{\circ}$)$\downarrow$ & 2.4 & 3.0 & \textbf{2.8} & 3.6 & \textbf{2.2} & 2.8 & \textbf{2.5} \\
    & $T_{e}$(mm)$\downarrow$ & 3.0 & 3.0 & \textbf{2.8} & 13.8 & \textbf{9.1} & 2.1 & \textbf{1.7} \\
    \hline
    \multirow{3}{*}{Bottle} & $5^{\circ}5$mm$\uparrow$ & 61.2 & 71.2 & \textbf{74.2} & 2.3 & \textbf{5.7} & 89.2 & \textbf{91.9} \\
    & $R_{e}$($^{\circ}$)$\downarrow$ & 5.6 & 3.7 & \textbf{3.6} & 12.0 & \textbf{5.2} & 2.4 & \textbf{2.1} \\
    & $T_{e}$(mm)$\downarrow$ & 5.0 & 2.9 & \textbf{2.7} & 16.1 & \textbf{10.3} & 2.0 & \textbf{1.8} \\
    \hline
    \multirow{3}{*}{Mug} & $5^{\circ}5$mm$\uparrow$ & 70.4 & 66.6 & \textbf{69.5} & 1.4 & \textbf{2.1} & 91.4 & \textbf{94.5} \\
    & $R_{e}$($^{\circ}$)$\downarrow$ & 4.6 & 4.3 & \textbf{4.0} & 9.6 & \textbf{6.9} & 2.3 & \textbf{2.1} \\
    & $T_{e}$(mm)$\downarrow$ & 3.8 & 2.9 & \textbf{2.6} & 13.0 & \textbf{9.1} & 1.9 & \textbf{1.6} \\
    \hline
    \multirow{3}{*}{Bowl} & $5^{\circ}5$mm$\uparrow$ & 58.6 & 58.7 & \textbf{62.0} & 0.2 & \textbf{0.3} & 69.6 & \textbf{75.8} \\
    & $R_{e}$($^{\circ}$)$\downarrow$ & 6.7 & 5.4 & \textbf{5.0} & 70.7 & \textbf{19.8} & 8.6 & \textbf{3.6} \\
    & $T_{e}$(mm)$\downarrow$ & 3.4 & 4.8 & \textbf{4.5} & 23.4 & \textbf{15.5} & 3.9 & \textbf{2.5} \\
    \hline
    \multirow{3}{*}{Overall} & $5^{\circ}5$mm$\uparrow$ & 70.0 & 67.9 & \textbf{70.8} & 2.0 & \textbf{3.0} & 84.1 & \textbf{87.5}  \\
    & $R_{e}$($^{\circ}$)$\downarrow$ & 4.4 & 4.1 & \textbf{3.9} & 21.19 & \textbf{8.1} & 3.8 & \textbf{2.6} \\
    & $T_{e}$(mm)$\downarrow$ & 3.7 & 3.4 & \textbf{3.2} & 16.23 & \textbf{11.1} & 2.5 & \textbf{1.9} \\
    \hline
\end{tabular}
\vspace{-0.3cm}
\label{tab:real_result}
\end{table}

\subsection{Ablation Studies}
\label{sec:5.5}

\textbf{Tactile Kinematic Cues.} To examine different tactile kinematic cues, we select real-world video clips containing at least one frame with slippage. Each video clip spans one second. Since two velocity predictors share the same output format, the kinematic states of the object could be fully obtained by either of them. We thus evaluate these two designs on the kinematic-only method and \model(ShapeAlign) and compare them with our proposed method. Table \ref{tab:tactile_ablation} shows the comparison. Compared with the method that ignores the object slippage and uses the optimization-based velocity predictor alone, leveraging a learning-based module to handle slippage cases could consistently achieve performance gain on both kinematics-only and multi-modal tracking approaches, while using such a learning-based module to fully replace optimization would significantly reduce the tracking effect. The reason is that contact points could indicate the movement of the sensor surface material with high precision hence its effect varies greatly depending on whether the object is slipping, on the contrary, the neural network could perform better under large object movement while obtaining noisier results in other cases. We thus combine the two designs as our proposed kinematic cues.

\begin{table}
    \centering
    \vspace{0.25cm}
    \caption{ABLATION STUDY ON TACTILE KINEMATIC CUES.}
    \vspace{-0.25cm}
    \begin{tabular}{|c|cc|c|c|c|}
        \hline
         \multirow{2}{*}{Method} & \multicolumn{2}{c|}{Designs} & \multirow{2}{*}{$5^{\circ}5$mm$\uparrow$} & \multirow{2}{*}{$R_{e}$($^{\circ}$)$\downarrow$} & \multirow{2}{*}{$T_{e}$(mm)$\downarrow$} \\
         & O$^{\rm 1}$ & L$^{\rm 2}$ & & & \\
         \hline
         \multirow{3}{*}{K} &  & \checkmark & 37.69 & 5.16 & 4.24 \\
         & \checkmark &  & 45.49 & 3.71 & 3.16 \\
         & \checkmark & \checkmark & \textbf{55.56} & \textbf{2.95} & \textbf{2.76} \\
         \hline
         \multirow{3}{*}{TE(SA)} &  & \checkmark & 47.05 & 3.51 & 2.43 \\
         & \checkmark &  & 56.26 & 2.75 & \textbf{2.21} \\
         & \checkmark & \checkmark & \textbf{59.87} & \textbf{2.67} & 2.24 \\
         \hline
    \end{tabular}
    \label{tab:tactile_ablation}
\vspace{0.05cm}
\footnotesize{$^{\rm 1}$Optimization-based VP (Velocity Predictor). $^{\rm 2}$Learning-based VP.}
\vspace{-0.2cm}
\end{table}

\textbf{Number of frames in Geometric-Kinematic Optimization.} While using \model, frame number $N$ for each geometric-kinematic optimization iteration determines the extent to use tactile signals. We evaluate \model(ShapeAlign) under different values of $N$ on real-world data and report the mean tracking errors among all categories in Table \ref{tab:diff_N}. Note that \model(ShapeAlign) degenerates into ShapeAlign when $N=1$.
The performance of \model(ShapeAlign) drops with either a small or a large $N$. A short sequence lacks adequate tactile information to improve vision-based object pose hypotheses, while a long one suffers from a cumulative error of object kinematic states.

\begin{table}
    \centering
    \caption{GEOMETRIC-KINEMATIC OPTIMIZATION WITH DIFFERENT FRAME NUMBER.}
    \vspace{-0.25cm}
    \begin{tabular}{|c|c|c|c|c|c|c|c|}
        \hline
         $N$ & 1 & 2 & 3 & 5 & 10 & 15 & 20 \\
         \hline
         $R_{e}$($^{\circ}$)$\downarrow$ & 4.10 & 4.03 & 3.99 & \textbf{3.87} & 3.88 & 3.91 & 3.96 \\
         \hline
         $T_{e}$(mm)$\downarrow$ & 3.39 & 3.29 & 3.26 & \textbf{3.19} & 3.21 & 3.27 & 3.38 \\
         \hline
    \end{tabular}
    \label{tab:diff_N}
\vspace{-0.5cm}
\end{table}

\subsection{Tracking Speed}
\label{sec:5.6}

Methods using BundleTrack are tested on a single NVIDIA RTX 2080Ti GPU, while others are tested on a single NVIDIA RTX 3090 GPU.
The speed of computing object kinematic states and producing geometric-kinematic optimization \textcolor{red}{is} 20 FPS (with $N$=5), which \textcolor{red}{is} faster than 11 FPS for ShapeAlign, 10 FPS for CAPTRA, and 1 FPS for BundleTrack, indicating that \model can improve the tracking performance for visual pose trackers with low additional time cost.

\subsection{Tracking Robustness}
\label{sec:5.7}

\begin{figure}[!tbp]
    \centering
    \vspace{0.25cm}
    \includegraphics[width=0.42\columnwidth]{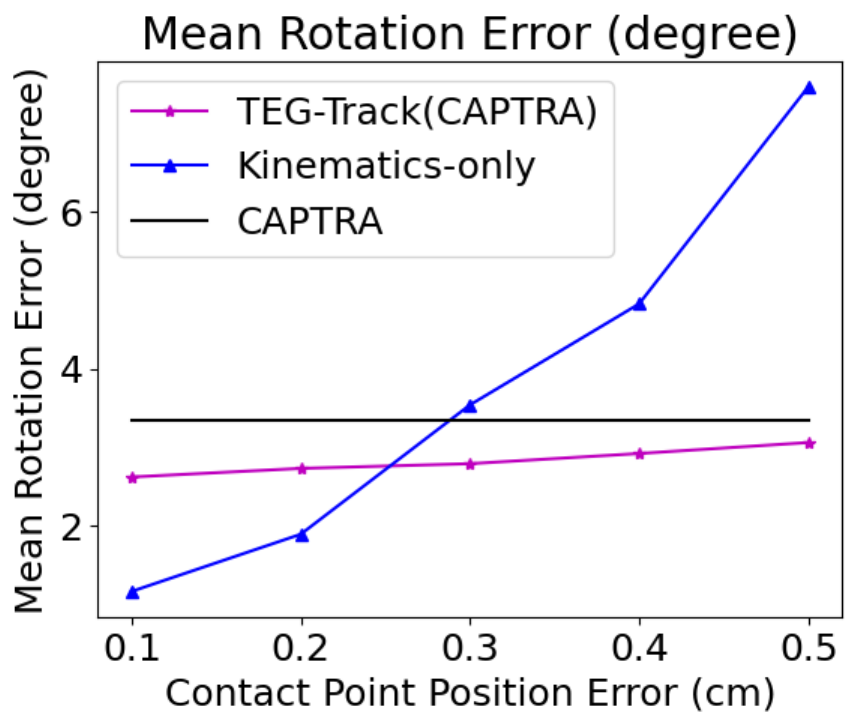}\hfill
    \includegraphics[width=0.42\columnwidth]{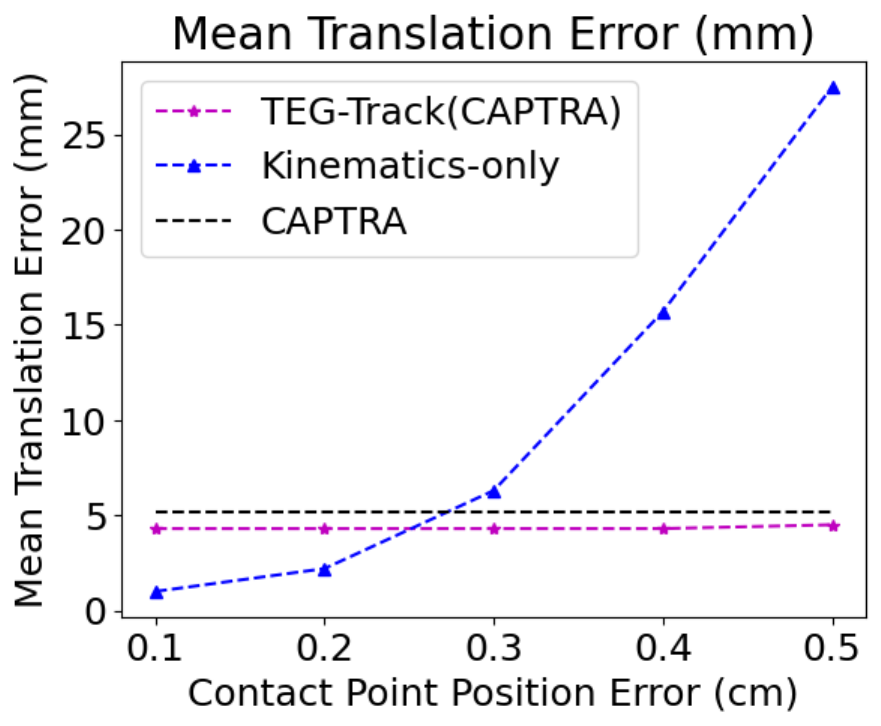}
\vspace{-0.25cm}
\caption{Mean rotation error ($^{\circ}$) and translation error (mm) under different tactile noise patterns. The horizontal axis indicates the standard deviation of the Gaussian noise added to the contact point positions. The solid and dashed lines show rotation and translation errors, respectively.}
\vspace{-0.6cm}
\label{fig:tracking_robustness_camera}

\end{figure}

The accuracy of object kinematic states hugely depends on the quality of marker flows detected by tactile sensors.
We evaluate \model under different kinematic noise patterns on synthetic data to test its robustness to various tactile sensing qualities. The kinematic noise contains the possible calibration error of tactile sensors for real scenarios and the detecting error of contact points. We select the \textit{camera} category as the evaluation set and evaluate the kinematics-only method, CAPTRA, and \model(CAPTRA) under different scales of Gaussian noise added on the position of contact points. As shown in Figure \ref{fig:tracking_robustness_camera}, \model(CAPTRA) is more robust to tactile sensor noise than the kinematics-only tracking method, meanwhile performing better than CAPTRA even under inaccurate tactile sensing.


\section{Limitations and Conclusion}
We explore the method leveraging visual and tactile sensing for generalizable in-hand object pose tracking. To this end, we present a novel framework \model with our core design to model kinematic cues for pose changes and integrate them with visual perception. We incorporate various visual pose trackers into \model and demonstrate consistent improvements on both synthetic and real-world data. We provide a visual-tactile in-hand object pose tracking dataset supporting relevant studies.

\textbf{Limitations.} One limitation is that our current method deals with only rigid objects but not articulated objects. 
We may need more complex grippers to support richer contacts with all parts of an articulated object.
Another limitation is that we currently do not design domain adaptation techniques for sim-to-real transfer. Multi-modal domain adaptation could also be an interesting future direction.




\bibliographystyle{IEEEtran}
\bibliography{ref}


\end{document}